\documentclass[journal]{IEEEtran}
\ifCLASSINFOpdf
  \usepackage[pdftex]{graphicx}
  \usepackage{booktabs}
  \usepackage{array} 
  \usepackage{url}
  \usepackage[numbers]{natbib}
  \usepackage{amsmath}
  \usepackage{makecell}
  \usepackage{multirow}

\else
\fi
\begin{document}
%
\title{Grade Guard: A Smart System for Short Answer Automated Grading}
%
%
%

\author{Niharika Dadu, Harsh Vardhan Singh, Romi Banerjee
\thanks{Niharika Dadu, Harsh Vardhan Singh, Romi Banerjee are with the Department of Computer Science and Engineering, Indian Institute of Technology (IIT), Jodhpur, India (e-mail: niharikadadu20@gmail.com; 
singh.187@iitj.ac.in;
romibanerjee@iitj.ac.in).}

}

\markboth{Journal of \LaTeX\ Class Files,~Vol.~14, No.~8, August~2015}%
{Shell \MakeLowercase{\textit{et al.}}: Bare Demo of IEEEtran.cls for IEEE Journals}
%



\maketitle

\begin{abstract}

The advent of large language models (LLMs) in the education sector has provided impetus to automate grading short answer questions. LLMs make evaluating short answers very efficient, thus addressing issues like staff shortage. However, in the task of Automated Short Answer Grading (ASAG), LLM responses are influenced by diverse perspectives in their training dataset, leading to inaccuracies in evaluating nuanced or partially correct answers. To address this challenge, we propose a novel framework, Grade Guard. 

\begin{itemize}
    \item To enhance the task-based specialization of the LLMs, the temperature parameter has been fine-tuned using Root Mean Square Error (RMSE).
    \item Unlike traditional approaches, LLMs in Grade Guard compute an Indecisiveness Score (IS) along with the grade to reflect uncertainty in predicted grades. 
    \item Introduced Confidence-Aware Loss (CAL) to generate an optimized Indecisiveness Score (IS).
    \item To improve reliability, self-reflection based on the optimized IS has been introduced into the framework, enabling human re-evaluation to minimize incorrect grade assignments.
\end{itemize}

Our experimentation shows that the best setting of Grade Guard outperforms traditional methods by $19.16\%$ RMSE in Upstage Solar Pro, $23.64\% $ RMSE in Upstage Solar Mini, $4.00$\% RMSE in Gemini $1.5$ Flash, and $10.20$\% RMSE in GPT $4$-o Mini. Future work includes improving interpretability by generating rationales for grades to enhance accuracy. Expanding benchmark datasets and annotating them with domain-specific nuances will enhance grading accuracy. Finally, analyzing feedback to enhance confidence in predicted grades, reduce biases, optimize grading criteria, and personalize learning while supporting multilingual grading systems will make the solution more accurate, adaptable, fair, and inclusive.



\end{abstract}


%
\IEEEpeerreviewmaketitle

\section{Introduction}

\IEEEPARstart{O}{ne} of the essential components of education comprises an assessment of quizzes, term papers, assignments, and exam papers to test students' learning and knowledge in the subject. These methodologies are ubiquitously used by educators to not only assess the level of a learner’s understanding but also help them devise future learning pathways, like updating the pace and structure of the course for the benefit of the student. In recent years, technology has been heavily integrated with educational systems and methodologies. For example, learning management systems like Moodle are widely used to distribute course materials, manage assignments, and facilitate communication. Additionally, adaptive learning platforms such as Coursera provide personalized learning experiences by tailoring content to individual student needs and progress. However, full integration has yet to be achieved, as tasks like evaluating term papers, assignments, and exam papers remain predominantly the responsibility of teachers and teaching assistants. This manual checking performed by teachers and teaching assistants is a time-consuming and cumbersome process. A 2021 UNESCO report highlights a shortage of over 1 million school teachers in India \cite{UNESCO2021}, exacerbated by rising higher education enrollments, financial constraints, and staff retirements, leading to greater workloads and higher student-to-teacher ratios. This is a global problem that many countries are currently facing, with the authors in \cite{UNESCO2024} addressing teacher shortages and transforming the profession, revealing an urgent need for 44 million primary and secondary teachers worldwide by 2030.  Introducing automated answer grading can alleviate the resource shortage by reducing staff workload and making the grading process more time-efficient \cite{aggarwal2024iunderstandigot}.



Quizzes, term papers, assignments, and exam papers contain a variety of question types ranging from multiple-choice questions (MCQs), multiple-select questions (MSQs), short answer questions, fill-in-the-blanks and many more. We emphasize the importance of non-binary grading in college exams, as subjective answers often exhibit partial correctness, and hence, we model the problem using regression-based loss in all subsequent experiments. S.~Burrows et al. in \cite{burrows2015eras} refer to a system that focuses on the content rather than the writing style to grade responses of length between one phrase and one paragraph to an objective question. 

In the last few years, LLMs have been introduced in the education sector. Its use is not only limited to students generating answers for assignments or exploiting its summarizing power to understand study materials like PowerPoint, notes, and research papers, but it is also used for evaluation purposes—developing automated systems for grading. Lee et al. \cite{lee2024college} have leveraged the ability of LLMs like ChatGPT and Gemini to understand the text and compare it with other texts to grade short answer-type questions. Xie et al. \cite{xie2024gradelikehumanrethinking} developed an ASAG system with grading rubrics based on questions and students' answers to grade answers more accurately, providing customized feedback based on the rubrics. However, ASAG systems have still not reached human grading capabilities. Developing ASAG systems requires ensuring dependability through fairness, uniformity, consistency, and accuracy \cite{Grevisse2024}. LLMs often assign varying scores to the same question-answer pair due to the large number of random variables, as noted by Lee et al. \cite{lee2024college}. This may lead to incorrect grade assignment. Developing LLM-based ASAG systems is further complicated due to limited benchmarked datasets for SAG in computer science, varying grading strictness among evaluators, and bias stemming from teachers' expectations of specific students. Furthermore, S.~Goyal et al. in \cite{goyal2024} discuss the risk of generating inappropriate content that can have legal implications. Thus, making automated grading systems challenging.

In this work, we have developed a framework, Grade Guard, described in Section 3 and Fig. \ref{fig:framework}, with the following contributions:
\begin{enumerate}
    \item Proposed a novel framework, Grade Guard, which aims to perform the task of automated question answering reliably by using large language models (LLMs). We outline each step in the framework, which includes prompt engineering, a creativity regulator system, confidence-aware loss (CAL) and indecisiveness regulator, and a self-reflective grader. 

    \item Engineered a new prompt that incorporates the context of the problem, assigns a role to the LLMs, and provides a baseline answer for one-shot learning. We have also added instructions for grading answers to reduce inconsistencies and subjectivity in grading. 
    
    
    \item Introduced a Creativity Regulator Module (CRM) to regulate creativity for LLM in ASAG by fine-tuning temperature on the sampled dataset using RMSE. 
    
    \item Introduced an Indecisiveness Regulator Module (IRM) that aims to reduce the error resulting from indecisiveness in the grading while maximizing the number of questions checked by LLM. This is done by introducing a new loss function—CAL—that combines RMSE with the penalty for indecisiveness, i.e., the instances where the LLM was indecisive and could not grade. 
    
    
    \item Introduced a self-reflective grader module (SRGM) that separates the grade generated by LLM based on the standard deviation in optimized Indecisive Score (IS). Grades with higher standard deviation are sent to the evaluator for checking.
    
    
\end{enumerate}

For a discussion on existing works, focusing on datasets and techniques relevant to ASAG, refer to Section 2. Section 3 details the proposed Grade Guard framework, including its methodology and experimental setup. In Section 4, we present the results and comparative analysis, highlighting the performance of Grade Guard against traditional LLMs. Finally, Section 5 outlines potential areas for improvement and future research directions.




\section{Related Works}
Several datasets have been published for benchmarking ASAG: Mohler et al. (2011) \cite{mohler2011} provided computer science questions with reference and student answers from the University of Texas; the SciEntsBank dataset \cite{Dzikovska2013} includes science questions graded on a 5-way categorical scale, and CU-NLP \cite{Tulu2021} features NLP question-answer pairs from Cukurova University. 

Initial works for ASAG tasks used classical approaches with the use of traditional techniques of NLP pertaining to feature extraction. Mohler et al. (2009) \cite{Mohler2009} used corpus-based semantic similarity along with lexical and syntactic features for an unsupervised technique for grading. A similar approach by Mihalcea et al. and Sahu et al. \cite{Mihalcea2006, Sahu2019} relies on deriving semantic overlap between the student answer and the model answer using semantic similarity measures, such as knowledge-based and corpus-based measures (e.g., WordNet), POS tagging, and inverse document frequency (IDF). 

ASAG, like other machine learning tasks, can benefit from transfer learning to address challenges arising from distribution changes in training and test data as discussed by S.~J.~Pan et al. in \cite{pan2010survey}. Later studies include the use of deep learning methods and advanced Natural Language Processing (NLP) techniques, such as the work by Bonthu et al. (2023) on automatic short answer grading using transfer learning and augmentation \cite{BONTHU2023106292}, and the study by Heilman and Madnani (2013) on domain adaptation and stacking for short answer scoring \cite{heilman-madnani-2013-ets}. Additionally, Jimenez et al. (2013) introduced SOFTCARDINALITY, a method for hierarchical text overlap in student response analysis \cite{jimenez2013softcardinality}. The introduction of transfer learning enabled architectures built on pre-trained embeddings like Embeddings from Language Models (ELMo) \cite{peters2018}, a deep bidirectional language model trained on a large text corpus. ELMo generates contextualized word representations that capture both syntax and semantics while modeling polysemy across linguistic contexts. These representations, derived from the model's internal states, can enhance existing models, significantly improving performance on NLP tasks. Similarly, Bidirectional Encoder Representations from Transformers (BERT) \cite{devlin2018bert} allowed models to be built on top of it by fine-tuning with just one additional output layer. The resulting pre-trained embeddings have represented the sentence space for grading answers through vector-similarity metrics. A comparative analysis of these models for ASAG-specific tasks can be found in \cite{gaddipati2020comparative}. Lavoie et al. \cite{lavoie2019latent} used Latent Semantic Analysis (LSA) as a similar approach by converting text into high-dimensional vectors. Fine-tuned variants of BERT, such as Sentence-BERT \cite{reimers2019sentencebert}, provided sentence-level embeddings that can directly be used to compare student and reference answers for testing correctness.

More recent attempts - \cite{lee2024college, aggarwal2024iunderstandigot, henkel2024largelanguagemodelsmake, duong2024automatic} have tested the performance of state-of-the-art LLMs for ASAG tasks. T. Duong et al. \cite{duong2024automatic} used pretrained embeddings provided by a pretrained LLM like OpenAI. In \cite{aggarwal2024iunderstandigot}, a first large-scale deployable system was proposed for IIT Bombay university exams in June $2024$, with a model evaluation mechanism composed of human verifiers. However, E. Pavlick et al. \cite{pavlick2023symbols} argued that LLMs lack symbolic structure and grounding and are unable to capture human language representation and understanding. This, coupled with other limitations such as hallucinations, suggests a necessity of ensuring the reliability of such systems before deployment. Building on the insights from existing works, the next section presents Grade Guard, a pipeline designed to enhance dependability in LLMs for the ASAG task.

 

\section{Methodology}

\begin{figure*}[h]
    \centering
    \includegraphics[width=1\textwidth]{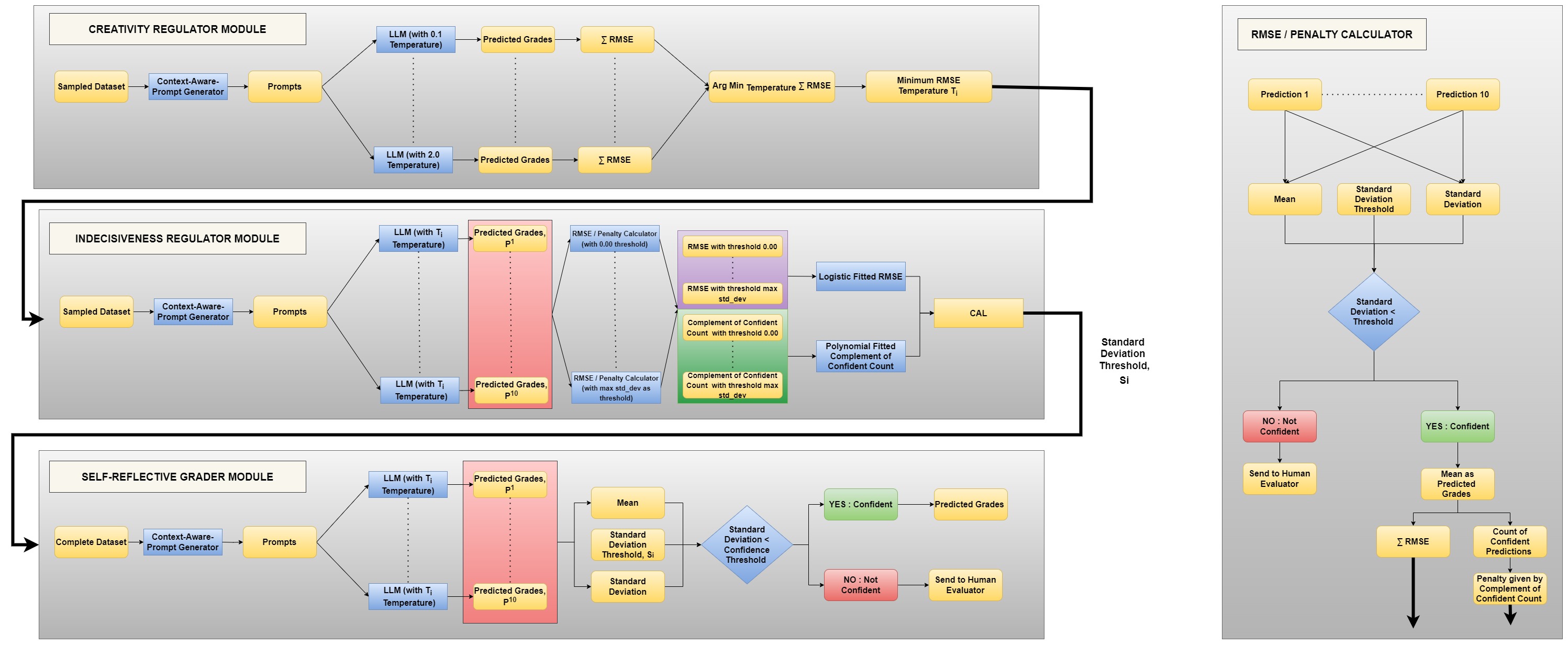} 
    \caption{Grade Guard Framework.}
    \label{fig:framework}
\end{figure*}

\subsection{Dataset Preparation}
We used the Mohler’s dataset \cite{mohler2011} comprising $87$ questions and $2,273$ responses from $31$ students in an introductory Computer Science course. Two independent evaluators graded responses on a $0$–$5$ scale, with the average serving as the final grade, rounded to the nearest $0.5$ for standardization.

\subsubsection{Cleaned dataset}
Student answers with tokenization issues due to special characters or missing spaces were removed to enhance dependability, resulting in the Mohler's cleaned dataset.


\subsubsection{Sampled dataset} 
Since LLM models are proprietary and costly to use, we fine-tuned the temperature parameter of the LLM and found the optimized IS on a sampled dataset. For creating this sampled dataset, we used a Score-Based Uniform Sampling (S-BUS) technique. For each question, we have divided the scores into $5$ ranges, i.e., [$0$,$1$], ($1$,$2$], ($2$,$3$], ($3$,$4$], ($4$,$5$]. We randomly sample one answer in each of these ranges for each question, which ensures the presence of answers with different levels of correctness for each question, allowing more generalization. Table \ref{tab:sbus} shows S-BUS for question id 1.6.

\begin{table}[h]
    \centering
    \caption{Score-Based Uniform Sampling (S-BUS) for Question ID 1.6}
    \renewcommand{\arraystretch}{1.3} 
    \begin{tabular}{|p{7cm}|c|}
        \hline
        \textbf{Student Answer} & \textbf{Grade} \\ 
        \hline
        After declaration of the variable's data type. & 1 \\ 
        \hline
        In the very beginning of the program, before the main starts. & 2 \\ 
        \hline
        Variables can be declared in classes and methods. & 2.5 \\ 
        \hline
       Anywhere in the same scope before they are used. & 3.5 \\ 
        \hline
        They can be declared globally just before the main method, but also outside of it. Variables can be subject to only the method they are scoped within, but would still be declared at the beginning of that method, inside of it. & 5 \\ 
        \hline
    \end{tabular}
    \label{tab:sbus}
\end{table}



\begin{figure}[h]
    \centering
    \includegraphics[width=0.6\columnwidth, height=5cm]{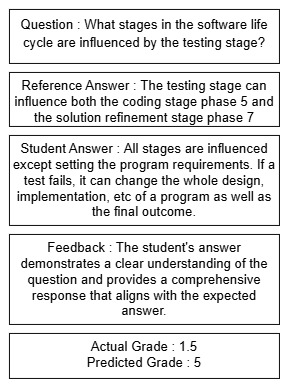} 
    \caption{Error in grade predicted by Upstage at 0.1 temperature.}
    \label{fig:wrongpred}
\end{figure}

\subsection{Grade Guard}

\begin{figure*}[h]
    \begin{minipage}[t]{0.45\textwidth}  
        \raggedright
        \includegraphics[height=6cm]{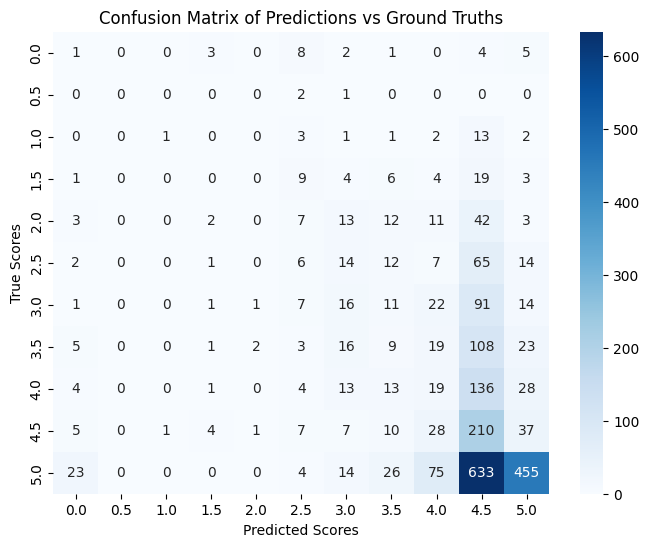} 
        \caption{Confusion Matrix for Predicted Grades vs True Grades : Upstage at 0.1 Temperature}
        \label{fig:Heatmap_upstage}
    \end{minipage}%
    \hspace{0.05\textwidth} 
    \begin{minipage}[t]{0.45\textwidth}  
        \raggedleft
        \includegraphics[height=6cm]{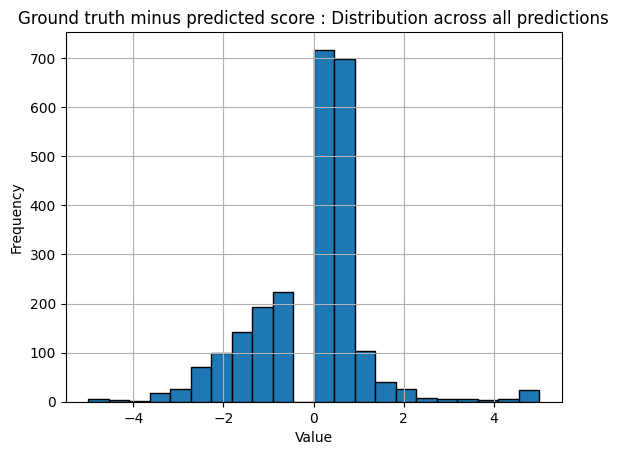} 
        \caption{Histogram of Error in Predictions : Upstage at 0.1 Temperature }
        \label{fig:barplot_upstage}
    \end{minipage}
\end{figure*}


Our framework, Grade Guard, consists of three key modules—CRM, IRM, and SRGM, as depicted in Fig. \ref{fig:framework}. It is designed to predict student grades based on a question statement, reference answer, and student answer from the Introduction to Computer Science course, using the Mohler cleaned dataset. This approach leverages the natural language understanding capabilities of large language models (LLMs) such as GPT-4-o-mini, Gemini 1.5-flash, Upstage solar-pro, and Upstage solar-1-mini-chat. However, we have observed that these LLMs, when used independently, often provide incorrect grades. A specific example is shown in Fig. \ref{fig:wrongpred}, while the incorrect predictions are highlighted along the non-diagonal elements in Fig. \ref{fig:Heatmap_upstage}. Frequently, the absolute error in grade prediction exceeds $2$, as illustrated in Fig. \ref{fig:barplot_upstage}. This makes the model unreliable for the evaluation process, as students who deserve a grade of $5$ may receive a $2$, or vice versa. To address this issue, Grade Guard introduces confidence measures designed to reduce such errors. The components of Grade Guard are discussed in the subsequent part:



\subsubsection{Prompt Engineered}
\begin{figure}[h]
    \centering
    \includegraphics[width=0.9\columnwidth]{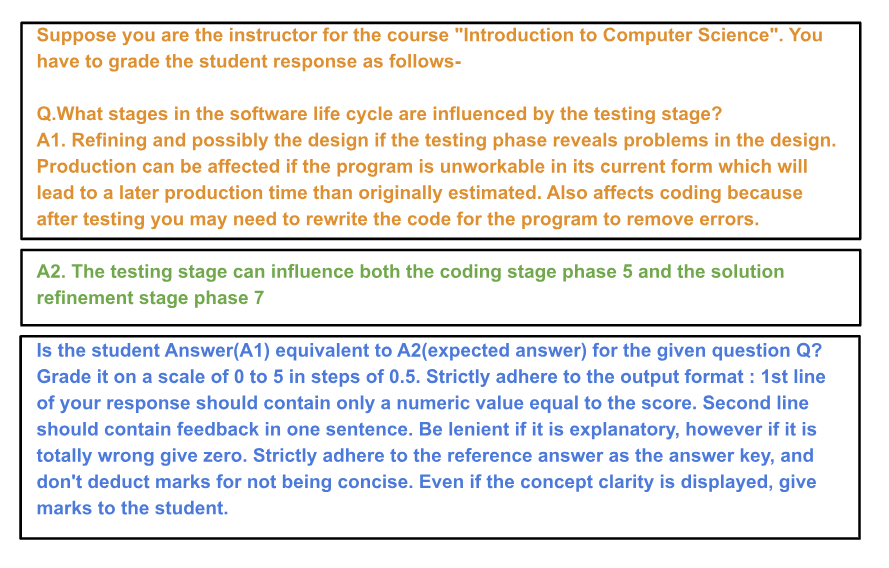} 
    \caption{Example of Context-Aware Prompt.}
    \label{fig:Prompt}
\end{figure}

To ensure that LLM understands the classification task at hand properly, we have created a Context-Aware-Prompt Generator using a $3$-step approach, described as follows: 

\begin{itemize}

\item Establish the context by assigning the LLM the role of an Introduction to Computer Science instructor and provide both the question and a reference answer to improve its understanding of the task.

\item Employed one-shot learning technique by giving one reference answer created by the instructor, which serves as a benchmark to compare against. 

\item Provided clear instructions for the grading policy and the output formats to ensure consistency in grading and allow easy grade retrieval from the outputs. An example of an engineered prompt has been illustrated in Fig. \ref{fig:Prompt}.

\end{itemize}

\subsubsection{Creativity Regulator Module (CRM)}
\begin{figure}[h]
    \centering
    \includegraphics[width=0.9\columnwidth]{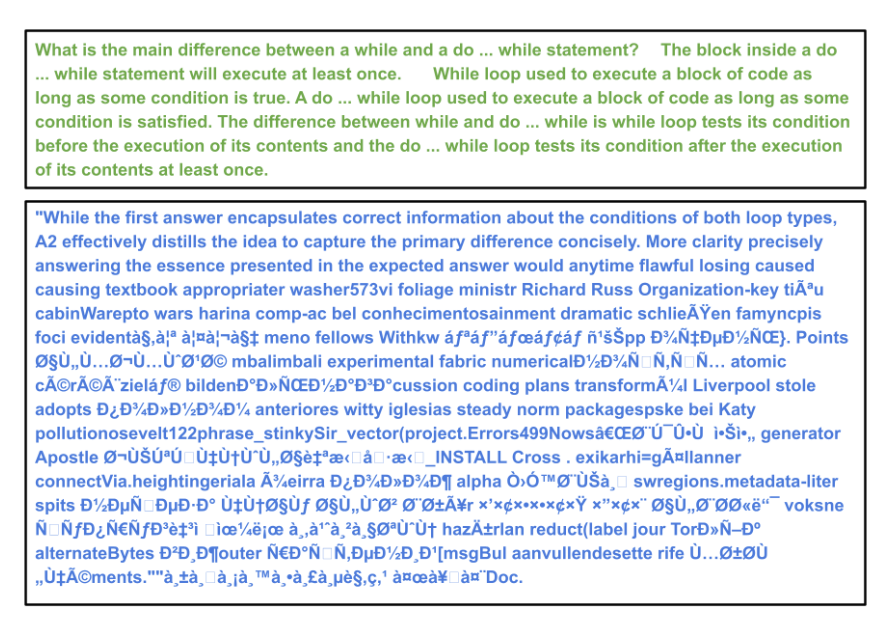} 
    \caption{Feedback Generated by GPT 4-o mini on temperature 1.9.}
    \label{fig:feedback_nonsense}
\end{figure}

Given the subjectivity of the problem, it is imperative to adjust the creativity levels. We processed the sampled dataset through the Context-Aware Prompt Generator to assess the impact of temperature variations on the grading process and used RMSE to evaluate model performance at different temperatures. Typically, the temperature for the LLMs varies from $0.0$ to $2.0$, controlling the level of randomness in the predictions. At temperatures close to $2.0$ some models like Gpt-40-mini tend to generate incoherent outputs as shown in Fig. \ref{fig:feedback_nonsense}.

\subsubsection{Indecisiveness Regulator Module (IRM) }

\begin{table}[h]
    \centering
    \caption{Variation in Grades predicted by Upstage Solar Pro at 0.7 Temperature}
    \small 
    \begin{tabular}{|>{\centering\arraybackslash}m{2.5cm}|>{\centering\arraybackslash}m{1.5cm}|>{\centering\arraybackslash}m{1.5cm}|>{\centering\arraybackslash}m{1.5cm}|} 
        \hline
        \textbf{Grade} & \textbf{Question 1} & \textbf{Question 2} & \textbf{Question 3} \\ 
        \hline
        Actual Grade         & 2.0   & 2.0   & 4.0   \\ \hline
        Predicted Grade 1    & 3.5   & 3.5   & 5.0   \\ \hline
        Predicted Grade 2    & 4.5   & 4.5   & 5.0   \\ \hline
        Predicted Grade 3    & 1.0   & 4.5   & 3.5   \\ \hline
        Predicted Grade 4    & 2.5   & 4.5   & 5.0   \\ \hline
        Predicted Grade 5    & 0.5   & 2.5   & 5.0   \\ \hline
        Predicted Grade 6    & 2.5   & 4.5   & 5.0   \\ \hline
        Predicted Grade 7    & 1.0   & 4.5   & 5.0   \\ \hline
        Predicted Grade 8    & 4.5   & 3.5   & 4.0   \\ \hline
        Predicted Grade 9    & 3.5   & 2.5   & 4.5   \\ \hline
        Predicted Grade 10   & 2.0   & 1.5   & 1.0   \\ \hline
        Standard Deviation   & 1.37  & 1.04  & 1.21  \\ \hline
    \end{tabular}
    \label{tab:grades_comparison}
\end{table}

\begin{figure*}[h]
    \centering  
    \begin{minipage}[t]{0.450\textwidth}  
        \centering
        \includegraphics[width=0.9\textwidth]{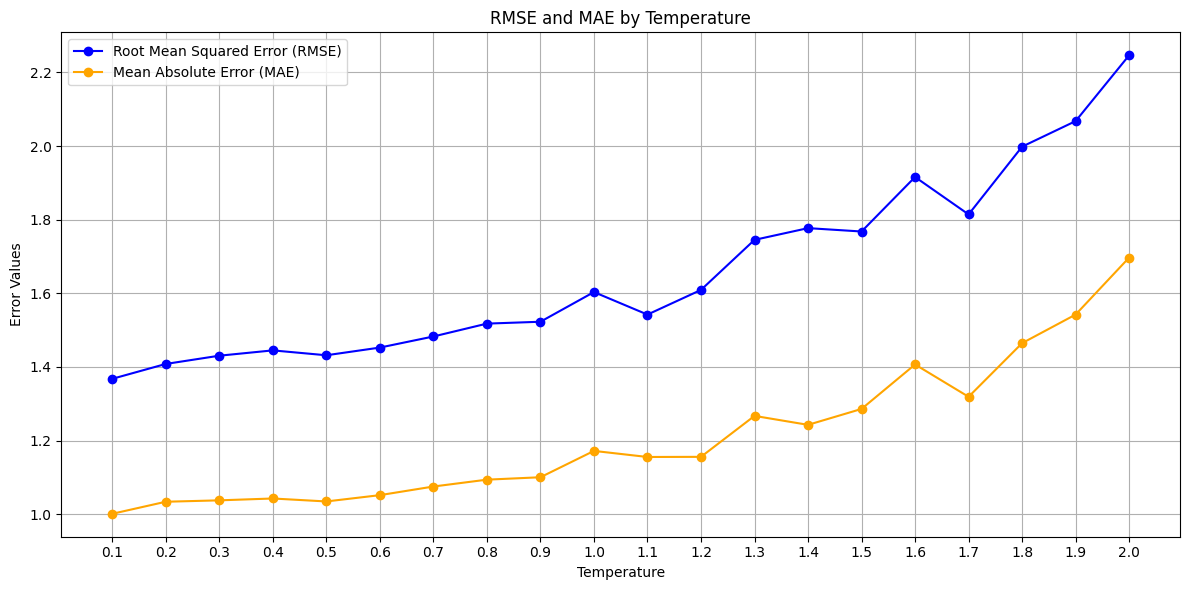}  
        \caption{Effect of Creativity on RMSE: Upstage Solar Pro}
        \label{fig:temp_var_solar_pro}
    \end{minipage}%
    \hspace{0.05\textwidth}  
    \begin{minipage}[t]{0.450\textwidth}  
        \centering
        \includegraphics[width=0.9\textwidth]{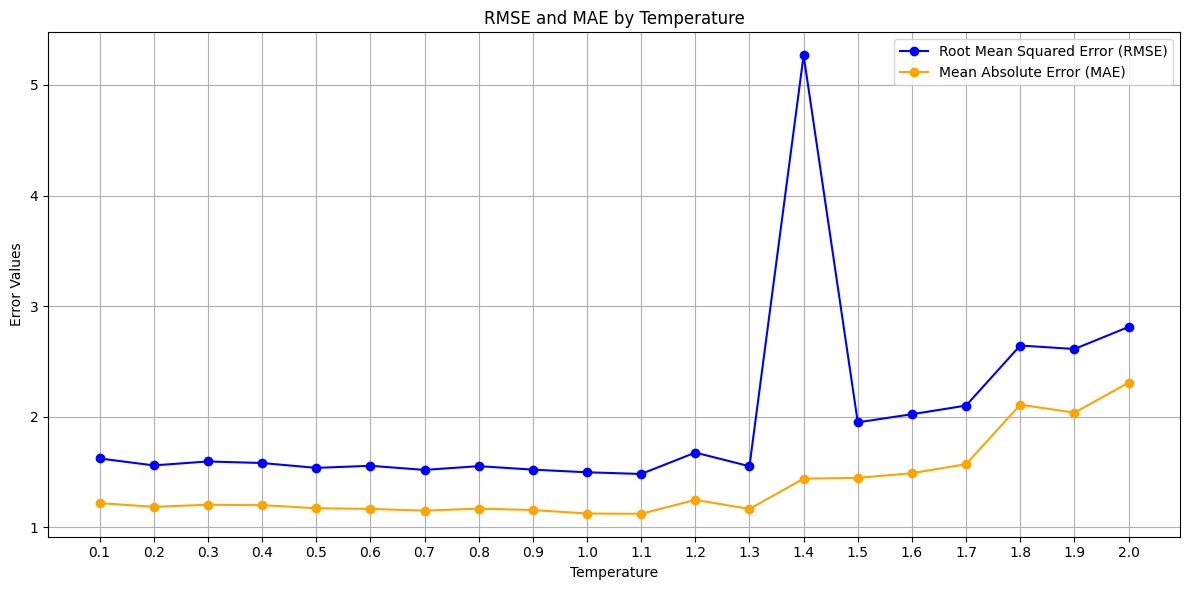}  
        \caption{Effect of Creativity on RMSE : Upstage Solar Mini}
        \label{fig:temp_var_solar_mini}
    \end{minipage}
\end{figure*}

\begin{figure*}[h]
    \centering  
    \begin{minipage}[t]{0.450\textwidth}  
        \centering
        \includegraphics[width=0.9\textwidth]{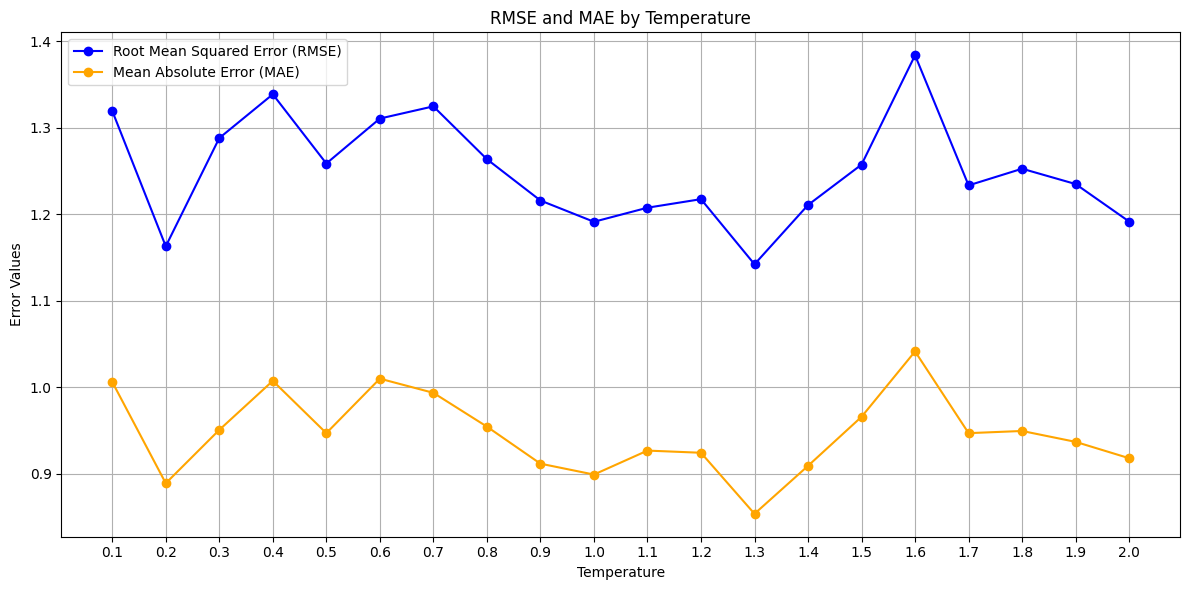}  
        \caption{Effect of Creativity on RMSE : Gemini 1.5-Flash}
        \label{fig:temp_var_gemini}
    \end{minipage}%
    \hspace{0.05\textwidth}  
    \begin{minipage}[t]{0.450\textwidth}  
        \centering
        \includegraphics[width=0.9\textwidth]{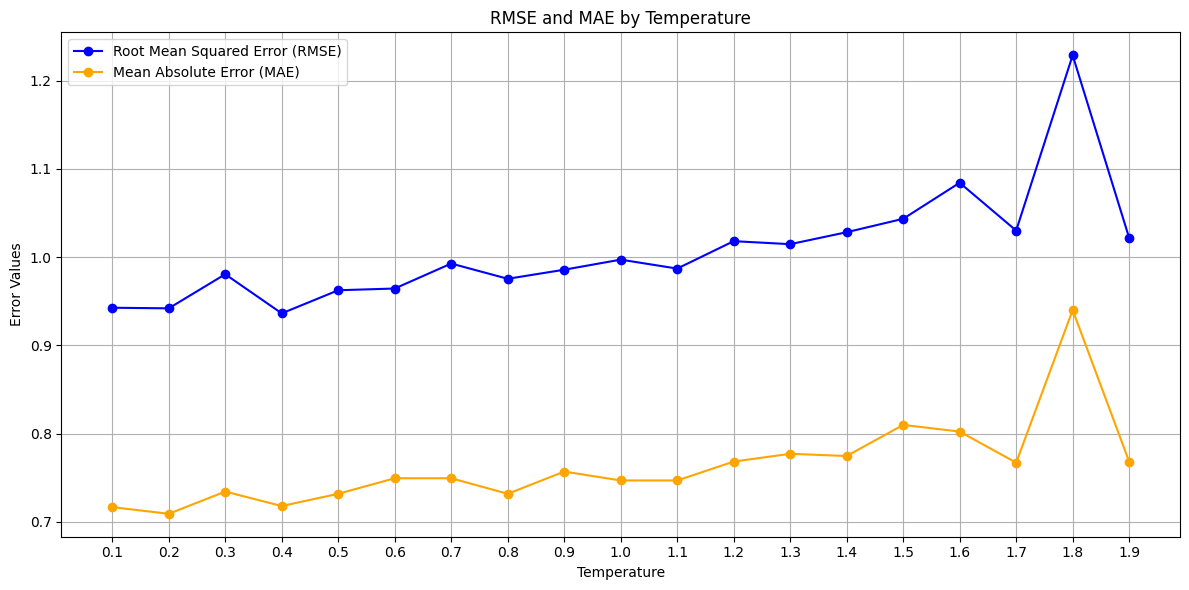}  
        \caption{Effect of Creativity on RMSE : GPT 4-o Mini}
        \label{fig:temp_var_gpt}
    \end{minipage}
\end{figure*}

Consider a triplet question $q_i$, reference answer $r_i$, and student answer $a_i$ for the sampled dataset; a prompt $p_i$ was generated. This prompt $p_i$ was sent to LLM 10 times, generating $10$ scores and feedbacks \(y_{i1}\), \(y_{i2}\), \dots \(y_{i10}\), and \(f_{i1}\), \(f_{i2}\), \dots \(f_{i10}\) respectively. This experimentation shows that repeatedly running the same prompt on the same model gave different grades with a very high standard deviation as shown in Table \ref{tab:grades_comparison}. 

A higher standard deviation suggests higher randomness in the predictions, or the model is uncertain of its predicted grade. To reduce this uncertainty, we developed the IRM. The mean of the 10 grades $\bar{y_i}$, generated for the same question-answer pair, serves as the final predicted grade as shown in Equation~\ref{eq:average_y}. Then the normalized standard deviation for the 10 scores $s_i$ is calculated as shown in Equation~\ref{eq:standard_dev} to determine the indecisiveness in the predictions. The normalized standard deviation, which we call Indecisiveness Score (IS) throughout the paper, serves as a measure of confidence. If the standard deviation, $s_i$, for the question-answer pair is higher than the selected threshold, $S_k$, the model is uncertain, and we send the question to be evaluated by a human. This is modeled by the function $\delta_i$ as shown in Equation~\ref{eq:standard_dev_condition}.

\begin{align}
\bar{y_i} = \frac{1}{t} \sum_{j=1}^{t} y_{ij} \label{eq:average_y}
\end{align}

\begin{align}
s_i = \frac{1}{10} \sqrt{\frac{1}{t-1}{\sum_{j=1}^{t} (y_{ij} - \bar{y_i})^2}} \label{eq:standard_dev}
\end{align}

\begin{align}
\delta_i = 
\begin{cases} 
      \text{1} & \text{if } s_i \leq S_k \\
      \text{0} & \text{if } s_i > S_k
\end{cases} \label{eq:standard_dev_condition}
\end{align}

 The challenge lies in finding the optimal threshold for IS. Smaller optimized IS will lead to less uncertainty in the grades assigned. However, setting this threshold to close to $0.0$ is counter-productive as it would lead to the re-evaluation of the majority of answers by the human, which is not desirable. To solve this, we have tried to minimize the RMSE and maximize the count of confident instances simultaneously, where the count of confident instances $N_k$ is given by Equation~\ref{eq:count}. For this, we have introduced Confidence-Aware Loss (CAL) with standardized and normalized scales. CAL combines the RMSE for the mean of $10$ predictions for the same question-answer pairs for which it is confident, $E_k$ as shown in Equation~\ref{eq:RMSE_conf}, with a penalty for the count of question-answer pairs where the model showed high variations in predictions, i.e., the model was not confident, $N'_k$ given by Equation~\ref{eq:penalty}. 

\begin{align}
N_k = \sum_{i=1}^{N} \delta_i \label{eq:count}
\end{align}

\begin{align}
E_k = \sqrt{\frac{1}{N_k} \sum_{i=1}^{N} \delta_i (y_i - \bar{y_i})^2} \label{eq:RMSE_conf}
\end{align}

\begin{align}
N'_k = 1 - \frac{N_k}{N} \label{eq:penalty}
\end{align}
 
 We obtain normalized RMSE, $E'_k$  as shown in Equation~\ref{eq:normalise_RMSE} for Normalized CAL (N-CAL) as well as standardized RMSE, $Z_{E_k}$ using Equations~\ref{eq:standardise_RMSE} for Standardized CAL (S-CAL). We have observed that the graph for RMSE with normalized standard deviation (excluding some very low normalized standard deviation points) is a logistic curve, whereas the complement of confidence count against normalized standard deviation is a polynomial fit. The logistic fitting for standardized RMSE, $f_1(Z_{E_k})$  and normalized RMSE, $f_2(E'_k)$ are given by Equation~\ref{eq:Standarse_Fit} and Equation~\ref{eq:normalise_fit} respectively. We observed that the $4$-degree polynomial curve $f_3(N'_k)$ best fits the penalty as shown by Equation~\ref{eq:poly_fit}.




\begin{align}
E'_k = \frac{E_k}{\max(E_k)} \label{eq:normalise_RMSE}
\end{align}

\begin{subequations}
\begin{align}
\mu_{E_k} = \frac{1}{n} \sum_{i=1}^{n} E_k \quad \text{and} \quad \sigma_{E_k} = \sqrt{\frac{1}{n-1} \sum_{i=1}^{n} (E_k - \mu_{E_k})^2}
\label{eq:standardise_RMSE_params}
\end{align}

\begin{align}
Z_{E_k} = \frac{E_k - \mu_{E_k}}{\sigma_{E_k}} \label{eq:standardise_RMSE}
\end{align} 
\end{subequations}


\begin{align}
f_1(Z_{E_k}) = \frac{L}{1 + e^{-k(Z_{E_k} - t_0)}} \label{eq:Standarse_Fit} 
\end{align}

\begin{align}
f_2(E'_k) = \frac{L}{1 + e^{-k(E'_k - t_0)}} \label{eq:normalise_fit}
\end{align}

\begin{align}
f_3(N'_k) = p_0 + p_1 N_k' + p_2 (N_k')^2 + p_3 (N_k')^3 + p_4 (N_k')^4 \label{eq:poly_fit}
\end{align}

We have then combined the standardized and normalized RMSE with a penalty for indecisiveness with equal weightage to obtain S-CAL and N-CAL, respectively, given by Equation~\ref{eq:SCAL} and Equation~\ref{eq:NCAL}.

\begin{align}
\text{S-CAL} = 0.5 \times f_1(Z_{E_k}) + 0.5 \times f_3(N'_k) \label{eq:SCAL}
\end{align}

\begin{align}
\text{N-CAL} = 0.5 \times f_2(E'_k) + 0.5 \times f_3(N'_k) \label{eq:NCAL}
\end{align}

\begin{figure*}[htbp]
    \centering
    \includegraphics[width=0.9\textwidth, height=5cm]{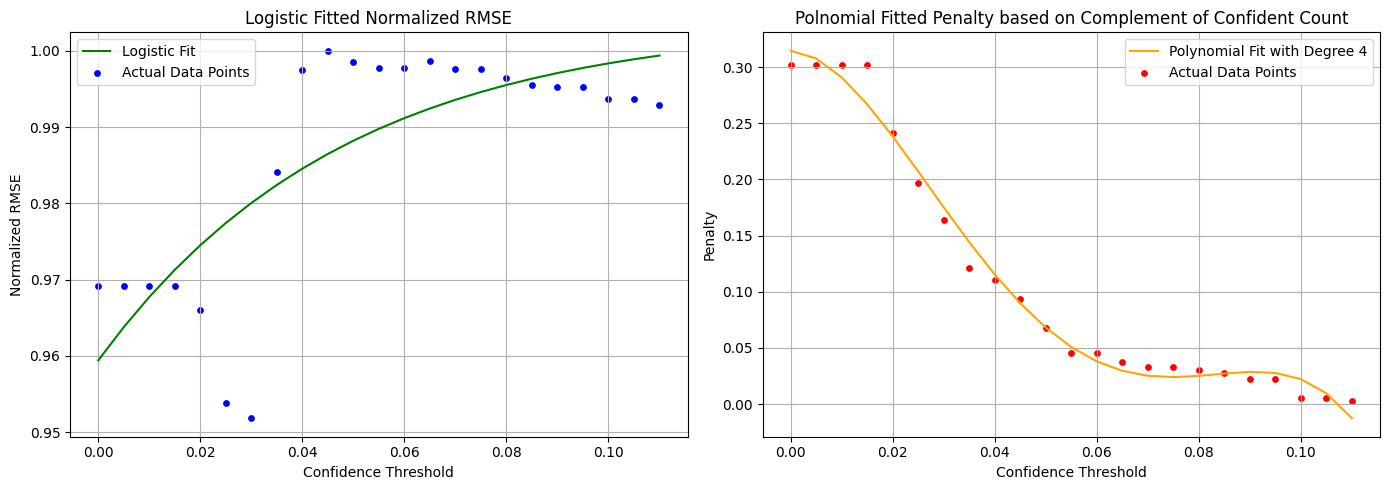}
    \caption{Logistic Fit for RMSE and Polynomial Fit for Penalty: Upstage Solar Pro at 0.1 Temperature}
    \label{fig:fitting_1}
\end{figure*}

\begin{figure*}[htbp]
    \centering
    \includegraphics[width=0.9\textwidth, height=5cm]{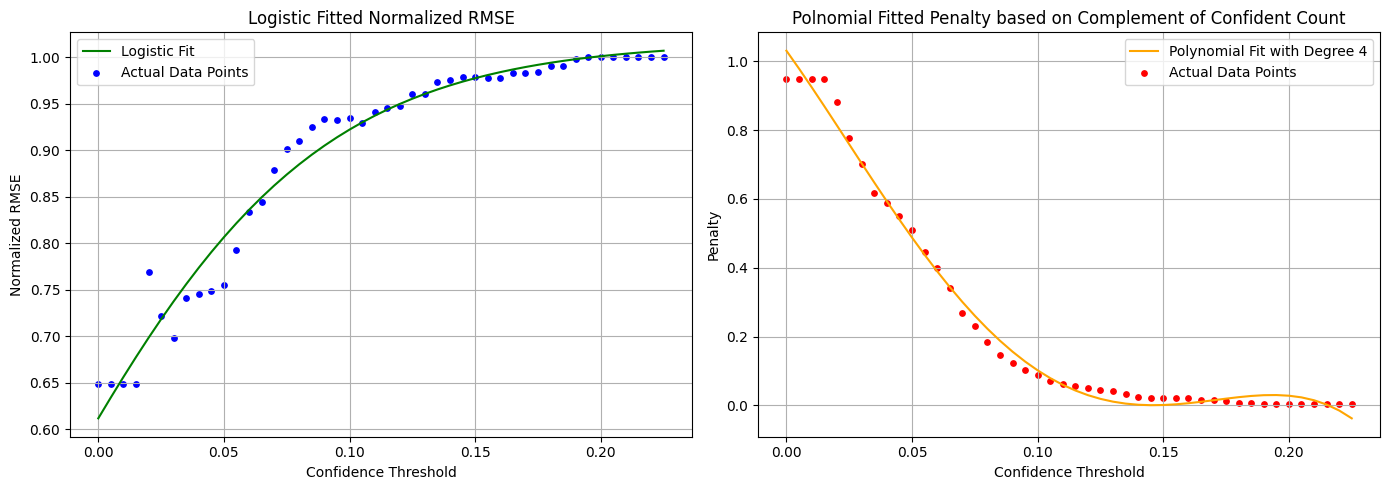}
    \caption{Logistic Fit for RMSE and Polynomial Fit for Penalty: Upstage Solar Mini at 1.1 Temperature}
    \label{fig:fitting_2}
\end{figure*}

\begin{figure*}[htbp]
    \centering
    \includegraphics[width=0.9\textwidth, height=5cm]{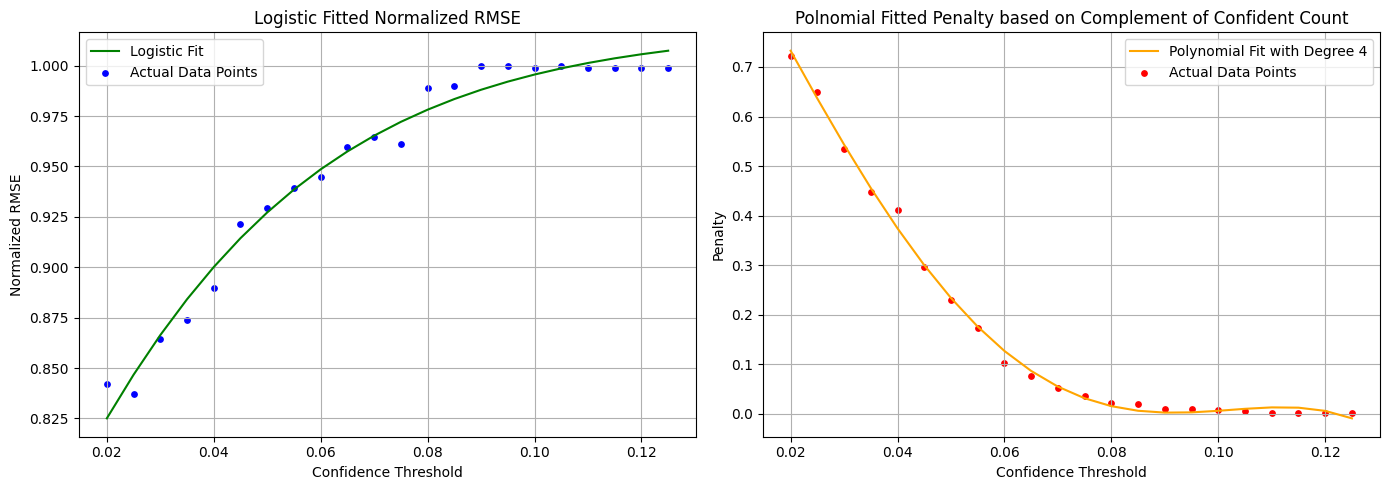}
    \caption{Logistic Fit for RMSE and Polynomial Fit for Penalty: Gemini 1.5-Flash at 1.3 Temperature}
    \label{fig:fitting_3}
\end{figure*}

\begin{figure*}[htbp]
    \centering
    \includegraphics[width=0.9\textwidth, height=5cm]{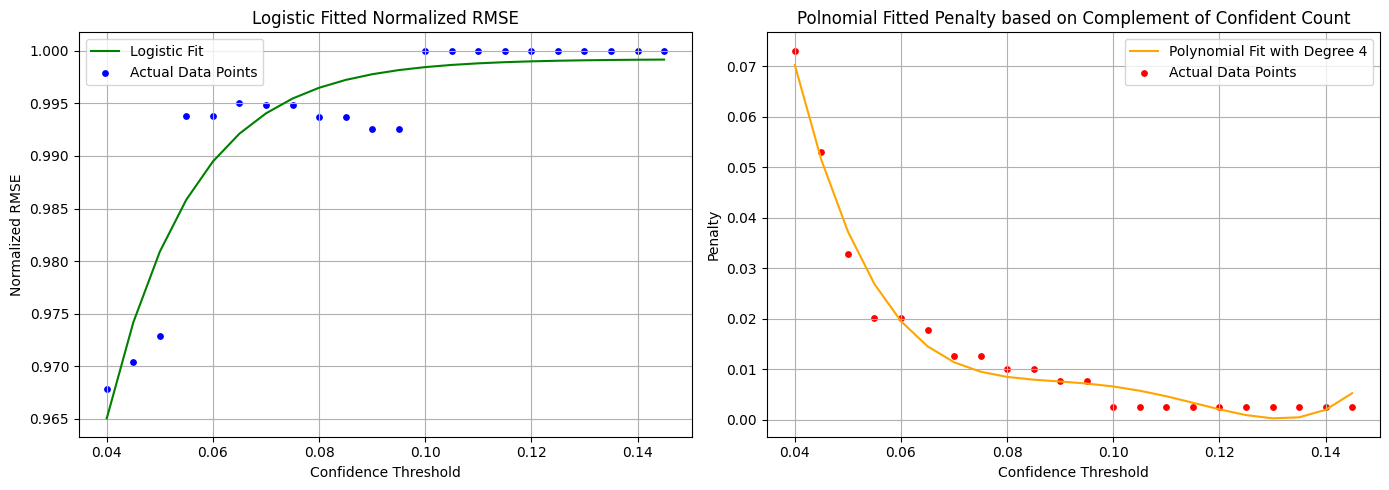}
    \caption{Logistic Fit for RMSE and Polynomial Fit for Penalty: GPT at 0.2 Temperature}
    \label{fig:fitting_gpt}
\end{figure*}

The first inflection point of the normalized CAL serves as the optimal threshold for the IS, as the inflection point determines the point where the penalty for non-confident instances starts to dominate the CAL instead of the RMSE. Alternatively, the standard deviation where the standardized-CAL is minimum or where CAL stops reducing was initially taken to be the optimal threshold for the confidence. It was observed that this approach resulted in a very high optimal confidence threshold for GPT and Gemini, as the minima is achieved for higher standard deviation, allowing even the predictions with high variability, which is not desirable.

\subsubsection{Self Reflective Grader Module (SRGM)} 

A context-aware prompt is generated using a triplet (a question, a reference answer, and a student answer) from the complete cleaned dataset and sent to the LLM $10$ times at the minimum RMSE temperature, producing $10$ scores and $10$ feedback responses. The normalized standard deviation of the $10$ scores, \( s_i \), is calculated, and based on this, self-reflection is introduced within the framework. If it is below the optimal IS, the model is deemed confident of its predictions. Conversely, if the normalized standard deviation exceeds this threshold, the model is considered to be uncertain, and the answer is forwarded to a human evaluator. In the next section, we present a detailed evaluation and analysis of each component of Grade Guard, comparing its performance to traditional LLMs.




\section{Results and Analysis}

\subsection{Creativity Regulator Results }
In this section, we evaluate the impact of creativity on LLM performance in grading the student answers from the sampled dataset. The creativity is regulated using the temperature of the model. As noted by A.~Agarwal et al. \cite{Agarwal2024}, higher temperatures are associated with enhanced creativity. However, how this creativity variation impacts the ASAG is not discussed. The key metrics used for this are Root Mean Square Error (RMSE) and Mean Absolute Error (MAE). Different models behave differently with variations in temperature. While for Upstage Solar Pro and GPT $4$-o Mini, increasing temperature leads to an increase in RMSE and MAE as illustrated in Fig. \ref{fig:temp_var_solar_pro} and Fig. \ref{fig:temp_var_gpt}, respectively, however, such clear-cut trend does not exist in Upstage Solar Mini and Gemini $1.5$ Flash as shown in Fig. \ref{fig:temp_var_solar_mini} and Fig. \ref{fig:temp_var_gemini}, respectively. Table \ref{tab:Min_rmse_temp} contains the experimentally determined minimum RMSE temperature for different models.

\begin{table}[h]
    \centering
    \caption{Min RMSE Temperature for Different Models}
    \begin{tabular}{|c|c|}
        \hline
        \textbf{Model} & \textbf{Temperature} \\ 
        \hline
        Upstage Solar Pro & 0.1 \\ \hline
        Upstage Solar Mini & 1.1 \\ \hline
        Gemini 1.5 Flash & 1.3 \\ \hline
        GPT 4-o Mini & 0.2 \\ \hline
    \end{tabular}
    \label{tab:Min_rmse_temp}
\end{table}

\subsection{CAL graphs and Confidence Thresholds}
In this section, we evaluate the confidence thresholds for LLM models, including Upstage Solar Pro, Upstage Solar Mini, Gemini $1.5$ Flash, and GPT $4$-o Mini.

\begin{figure*}[htbp]
    \centering
    \includegraphics[width=1\textwidth, height=5cm]{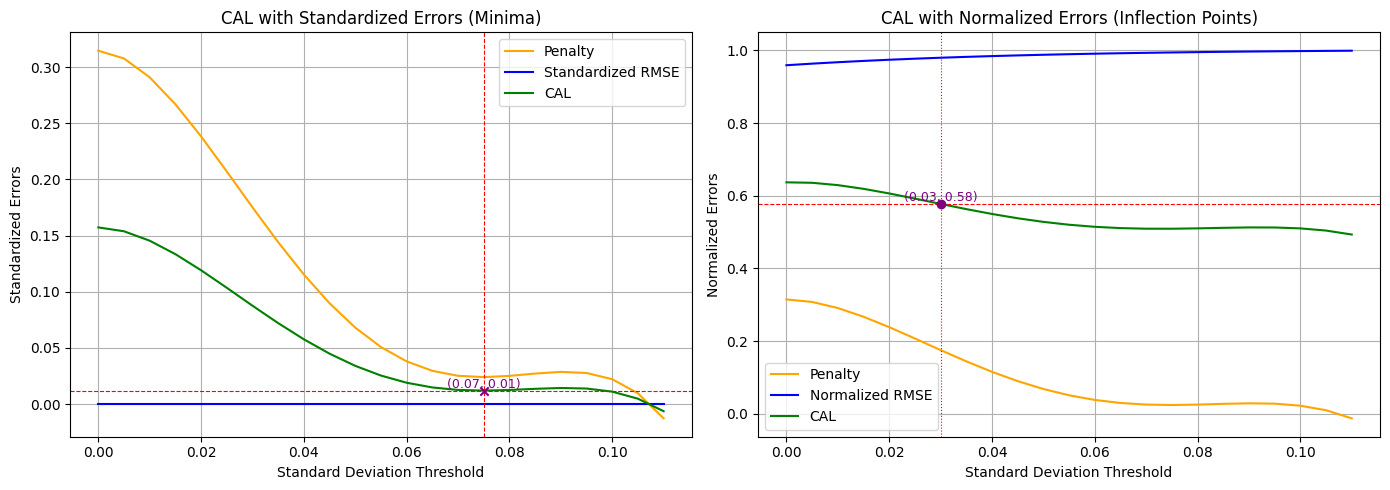}
    \caption{CAL for Upstage Solar Pro at 0.1 Temperature Using Minima as optimal Confidence Threshold}
    \label{fig:CAL_solar_pro}
\end{figure*}

\begin{figure*}[htbp]
    \centering
    \includegraphics[width=1\textwidth, height=5cm]{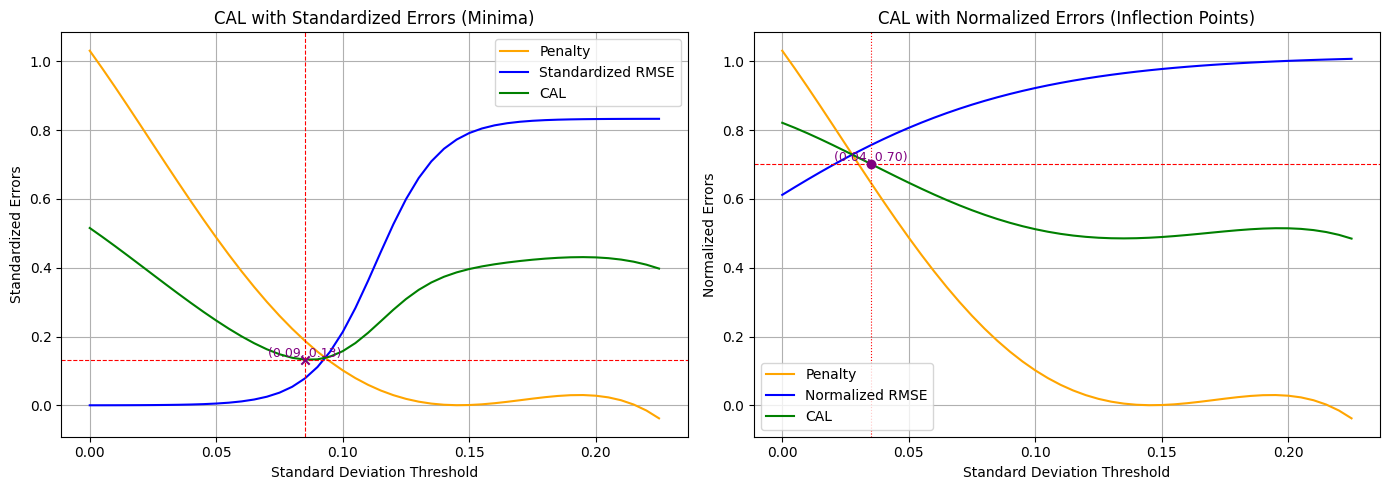}
    \caption{CAL for Upstage Solar Mini at 1.1 Temperature Using Minima as optimal Confidence Threshold}
    \label{fig:CAL_solar_mini}
\end{figure*}

\begin{figure*}[htbp]
    \centering
    \includegraphics[width=1\textwidth, height=5cm]{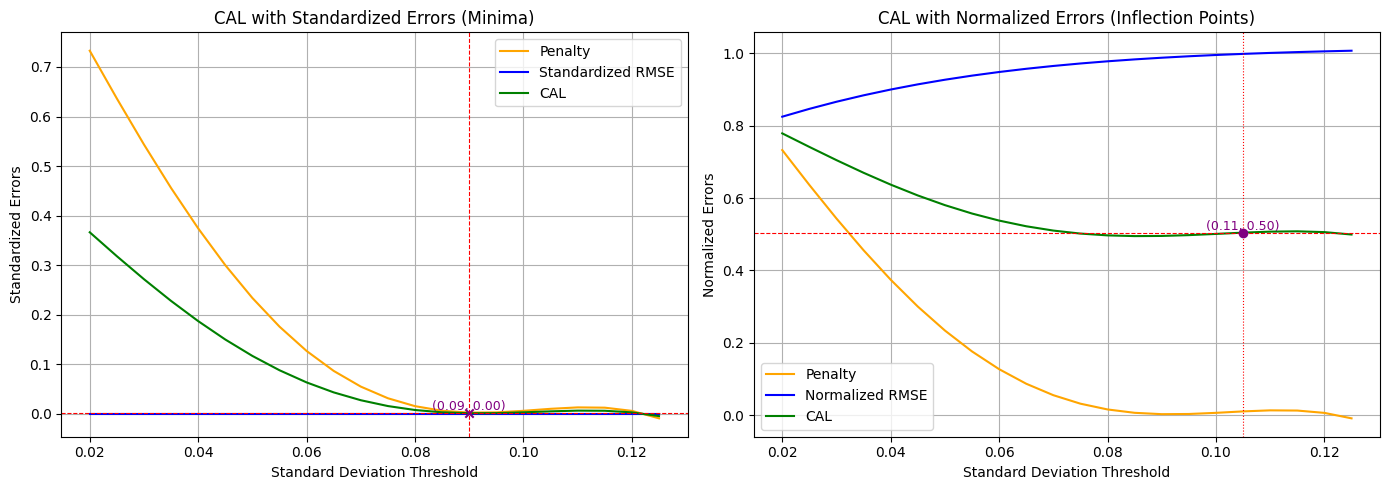}
    \caption{CAL for Gemini 1.5-Flash at 1.3 Temperature Using Minima as optimal Confidence Threshold}
    \label{fig:CAL_gemini}
\end{figure*}

\begin{figure*}[htbp]
    \centering
    \includegraphics[width=1\textwidth, height=5cm]{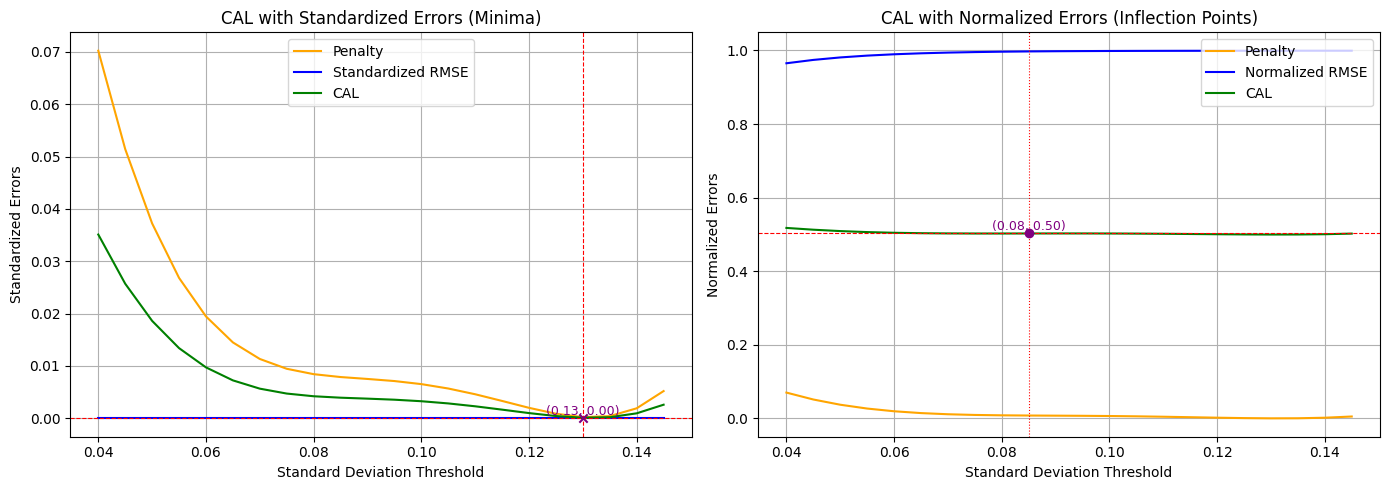}
    \caption{CAL for GPT 4-o mini at 0.2 Temperature Using Minima as optimal Confidence Threshold}
    \label{fig:CAL_gpt}
\end{figure*}

\begin{itemize}
\item Fig. \ref{fig:fitting_1}-\ref{fig:fitting_gpt} display the polynomial fit for the penalty and the logistic fit for RMSE as they vary with the confidence threshold for the models at a minimum RMSE temperature. 
\item The RMSE vs. normalized standard deviation, excluding a few very low standard deviation points (like 0.01), is observed to be logistic for all the models. Thus we have excluded these points in fitting the logistic curves. A small standard deviation for the optimal IS is counterproductive, as it would classify the majority of predicted grades as indecisive, undermining the model's effectiveness.

\item The fitted RMSE and penalty are then combined to generate the S-CAL and N-CAL, as illustrated in Fig. \ref{fig:CAL_solar_pro}-\ref{fig:CAL_gpt} for Upstage Solar Pro, Upstage Solar Mini, Gemini $1.5$ Flash, and GPT $4$-o Mini, respectively.

\item S-CAL uses the minima as the optimal IS, whereas for N-CAL it is the inflection point. These minima are highlighted in the figures that are compiled in Table \ref{tab:optimal_is}. 

\item S-CAL gives the optimal value at the relatively higher standard deviation, which is undesirable as it allows grades with higher indecisiveness to be assigned, which should ideally be rechecked by humans.

\item Using N-CAL, all the models under consideration except the Gemini have relatively lower optimal values, which is desired. 

\end{itemize}

\begin{table}[h]
    \centering
    \caption{Optimal Indecisiveness Score for Different Models}
    \begin{tabular}{|c|c|c|} 
        \hline
        \textbf{Model} & \textbf{\makecell{S-CAL Optimal\\ Indecisiveness Score}} & \textbf{\makecell{N-CAL Optimal\\ Indecisiveness Score}} \\ 
        \hline
        Upstage Solar Pro & 0.07 & 0.03  \\ \hline
        Upstage Solar Mini & 0.09 & 0.04  \\ \hline
        Gemini 1.5 Flash & 0.09 & 0.11 \\ \hline
        GPT 4-o Mini & 0.13 & 0.08 \\ \hline
    \end{tabular}
    \label{tab:optimal_is}
\end{table}





\begin{table*}[htbp]
    \centering
    \caption{Performance of Different Models with Various Settings at min RMSE temperature}
    \begin{tabular}{|c|c|c|c|c|c|c|c|}
        \hline
        Model Name & Ranges & \begin{tabular}[c]{@{}c@{}}Misclassifications without \\Grade Guard \end{tabular} &
        \multicolumn{2}{c|}{\begin{tabular}[c]{@{}c@{}}Misclassifications with \\Grade Guard\end{tabular}} & 
        \multicolumn{2}{c|}{\begin{tabular}[c]{@{}c@{}}Reduction in misclassified points\\ \end{tabular}} \\ \hline
        & & & \begin{tabular}[c]{@{}c@{}}Inflection Point\end{tabular} & \begin{tabular}[c]{@{}c@{}}Minima\end{tabular} & \begin{tabular}[c]{@{}c@{}}Inflection Point\end{tabular} & \begin{tabular}[c]{@{}c@{}}Minima\end{tabular} \\ \hline

      \multirow{4}{*}{Upstage Solar Pro} & [-5,-1)  & 214 & 178 & 207 & 16.82\% & 3.27\% \\ \cline{2-7}
                                        & [-1,1]-\{0\}  & 744 & 611 & 722 & 17.87\% & 2.95\% \\ \cline{2-7}
                                        & \{0\}  & 1198 & 1160 & 1190 & 3.17\% & 0.67\% \\ \cline{2-7}
                                        & (1,5]  & 264 & 136 & 219 & 48.49\% & 17.04\% \\ \hline
      \multirow{4}{*}{Upstage Solar Mini} & [-5,-1)  & 364 & 109 & 297 & 70.05\% & 18.40\% \\ \cline{2-7}
                                        & [-1,1]-\{0\}  & 1104 & 677 & 1033 & 38.67\% & 6.43\% \\ \cline{2-7}
                                        & \{0\}  & 714 & 546 & 686 & 23.53\% & 3.92\% \\ \cline{2-7}
                                        & (1,5]  & 238 & 109 & 208 & 54.20\% & 12.60\% \\ \hline
      \multirow{4}{*}{Gemini 1.5 Flash} & [-5,-1)  & 77 & 77 & 77 & 0.0\% & 0.0\% \\ \cline{2-7}
                                        & [-1,1]-\{0\}  & 1334 & 1333 & 1329 & 0.07\% & 0.37\% \\ \cline{2-7}
                                        & \{0\}  & 331 & 330 & 328 & 0.30\% & 0.90\% \\ \cline{2-7}
                                        & (1,5]  & 678 & 676 & 662 & 0.30\% & 2.36\% \\ \hline
      \multirow{4}{*}{GPT 4-o mini} & [-5,-1)  & 73 & 71 & 73 & 2.74\% & 0.0\% \\ \cline{2-7}
                                     & [-1,1]-\{0\}  & 1034 & 1033 & 1034 & 0.01\% & 0.0\% \\ \cline{2-7}
                                     & \{0\}  & 583 & 578 & 579 & 0.86\% & 0.68\% \\ \cline{2-7}
                                     & (1,5]  & 730 & 717 & 725 & 1.78\% & 0.69\% \\ \hline
    \end{tabular}
    \label{tab:misclassification_summary_all_models}
\end{table*}

 \begin{table*}[htbp]
    \centering
    \caption{RMSE and Confident Instances Summary For Various Grade Guard Settings}
    \begin{tabular}{|c|c|c|c|c|c|c|}
        \hline
        Model & 
        \multicolumn{2}{c|}{\begin{tabular}[c]{@{}c@{}}RMSE without Grade Guard\end{tabular}} & 
        \multicolumn{2}{c|}{\begin{tabular}[c]{@{}c@{}}RMSE with Grade Guard\end{tabular}} & 
         \multicolumn{2}{c|}{\begin{tabular}[c]{@{}c@{}} Confident Instances Count \\with Grade Guard\end{tabular}} \\ \hline

         & 
        \begin{tabular}[c]{@{}c@{}}Default temperature\end{tabular} & 
        \begin{tabular}[c]{@{}c@{}} min RMSE  temperature\end{tabular} & 
        \begin{tabular}[c]{@{}c@{}}  Inflection point\end{tabular} & 
        \begin{tabular}[c]{@{}c@{}}  Minima\end{tabular} & 
        \begin{tabular}[c]{@{}c@{}}  Inflection point\end{tabular} & 
        \begin{tabular}[c]{@{}c@{}}  Minima\end{tabular} 
       \\ \hline
        
    Upstage Solar Pro & 1.20 & 1.08 & 0.97 & 1.00 & 2085 & 2338\\ \hline
   Upstage Solar Mini & 1.10 & 1.12 & 0.84 & 0.95 & 1441 & 2224 \\ \hline
     Gemini 1.5 Flash & 1.00 & 1.02 & 0.96 & 0.95 & 2416 & 2396 \\ \hline
         Gpt 4-o Mini & 0.98 & 0.93 & 0.88 & 0.89 & 2399 & 2411 \\ \hline
    \end{tabular}
    \label{tab:performance_summary}
\end{table*}

\subsection{LLMs performance without Grade Guard}
Our experimentation showed that LLMs graded answers in the Mohler's cleaned dataset with an absolute difference greater than $1$. Specifically, in Upstage Solar Pro, $19.75\%$ of the predicted grades exhibited an absolute error exceeding $1$, while this value increased to $33.18\%$ for GPT $4$-o Mini at their minimum RMSE temperatures. For Upstage Solar Mini, it was $24.87\%$ of its predicted grades falling into this category. 

\subsection{LLMs performance with Grade Guard}
LLMs performance improved after using Grade Guard. The RMSE decreased for all the models considered, as shown in Table \ref{tab:performance_summary}. This decrease in RMSE can be attributed to the reduction in misclassification of grades with errors greater than $1$ and less than $-1$ and using the mean of the $10$ grades generated for the same question-answer pair. Grade Guard Framework gives a $19.16\%$ reduction in RMSE for Upstage Solar Pro. This reduction is higher for Upstage Solar Mini, which is $23.64\%$. For Gemini $1.5$ Flash and GPT $4$-o Mini, the reductions are $4.00\%$ and $10.20\%$, respectively. In Upstage Solar Pro, the amount of misclassifications decreased significantly with a $16.82\%$, $17.87\%$, and $48.49\%$ decrease in the $[-5,-1)$, $[-1,1]-\{0\}$, and $(1,5]$ ranges, respectively, as shown in Table \ref{tab:misclassification_summary_all_models}. This effect is more pronounced in Upstage Solar Mini, where there is a reduction of $54.20\%$, $38.67\%$, and $70.05\%$ in the $[-5,-1)$, $[-1,1]-\{0\}$, and $(1,5]$ ranges, respectively. Due to larger optimal values, the decreases for the GPT and Gemini models are not that pronounced. We discuss possible enhancements and suggest future research directions in the following section.

\section{Discussion}
We have identified 4 areas of improvement. Firstly, the models are not open-sourced, and we cannot look at the Bag of Words created by the LLM corresponding to each of the question-answer pairs. In the future, self-reflection can be added to the Bag-Of-Words generated. Second, LLMs are priced based on the input and output tokens, making grading a large number of students and questions costly. Third, few benchmarked datasets are available for ASAG. This problem is heightened by the fourth challenge, which is domain specificity. Less and less datasets are available for the domain-specific datasets like Introduction to Computer Science, Operating Systems, etc. Developing the models that will be able to transfer the learning from one domain to another remains an important research task to save resources and time.

\section{Conclusion}
This work presents a novel and reliable architecture for automated grading for short answers. Our proposed architecture performs competitively on the Mohler Dataset for four different types of LLMs, highlighting the importance of temperature-based tuning and confidence-based thresholds. We further show that adding a self-reflection unit helps in efficiently tackling inaccuracies in the grades generated by LLM. Future work includes exploring the use of these architectures for other forms of questions and exploring the use of reinforcement learning to further improve the prompt and model performance. Additionally, combining the strengths of different LLMs through an ensemble model could further improve confidence and accuracy in ASAG.

\ifCLASSOPTIONcaptionsoff
  \newpage
\fi

\end{document}